\documentclass[10pt,twocolumn,letterpaper]{article}

\usepackage{cvpr}              % To produce the CAMERA-READY version
\usepackage{graphicx}
\usepackage{amsmath}
\usepackage{amssymb}
\usepackage{booktabs}
\usepackage[accsupp]{axessibility}

\usepackage{mathtools}
\usepackage{bm}
\usepackage{bbm}
\usepackage[rgb,dvipsnames]{xcolor}
\usepackage{tikz}
\usepackage{pgfplots}
\pgfplotsset{compat=1.17}
\usepackage{threeparttable}
\usepackage{multirow}
\usepackage{pifont}
\newcommand{\cmark}{\ding{51}}
\newcommand{\xmark}{\ding{55}}

\usepackage[pagebackref,breaklinks,colorlinks]{hyperref}

\usepackage[capitalize]{cleveref}
\crefname{section}{Sec.}{Secs.}
\Crefname{section}{Section}{Sections}
\Crefname{table}{Table}{Tables}
\crefname{table}{Tab.}{Tabs.}

\def\cvprPaperID{2106} % *** Enter the CVPR Paper ID here
\def\confName{CVPR}
\def\confYear{2023}

\begin{document}

\title{Efficient Movie Scene Detection using State-Space Transformers}

% \and
\author{Md Mohaiminul Islam$^{1}\thanks{Research done while MI was an intern at Comcast Labs.}$
\quad
Mahmudul Hasan$^2$
\quad
Kishan Shamsundar Athrey$^2$\\
Tony Braskich$^2$
\quad\quad
Gedas Bertasius$^1$\\
$^1$UNC Chapel Hill \quad\quad $^2$Comcast Labs
}
\maketitle

%%%%%%%%% ABSTRACT
\begin{abstract}
   The ability to distinguish between different movie scenes is critical for understanding the storyline of a movie. However, accurately detecting movie scenes is often challenging as it requires the ability to reason over very long movie segments. This contrasts with most existing video recognition models, which are typically designed for short-range video analysis. This work proposes a State-Space Transformer model that can efficiently capture dependencies in long movie videos for accurate movie scene detection. Our model, called TranS4mer, is built using a novel S4A building block, combining the strengths of structured state-space sequence (S4) and self-attention (A) layers. Given a sequence of frames divided into movie shots (uninterrupted periods where the camera position does not change), the S4A block first applies self-attention to capture short-range intra-shot dependencies. Afterward, the state-space operation in the S4A block aggregates long-range inter-shot cues. The final TranS4mer model, which can be trained end-to-end, is obtained by stacking the S4A blocks one after the other multiple times. Our proposed TranS4mer outperforms all prior methods in three movie scene detection datasets, including MovieNet, BBC, and OVSD, while being $2\times$ faster and requiring $3\times$ less GPU memory than standard Transformer models. We will release our code and models.
\end{abstract}

% \vspace{-0.3cm}
\section{Introduction}
\label{sec:intro}

\begin{figure*}[t]
    \vspace{-5mm}
  \centering
    \includegraphics[width=1\linewidth]{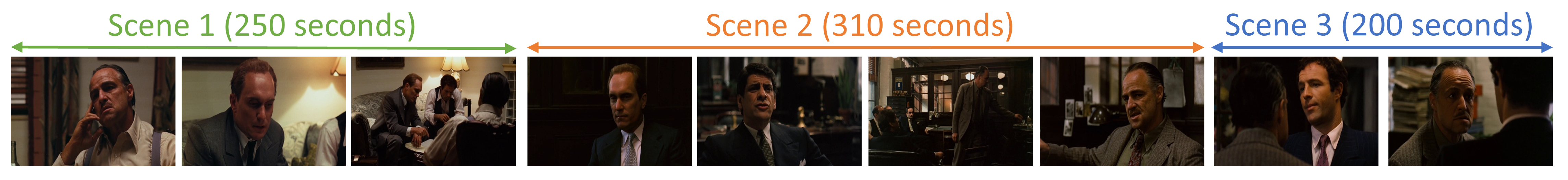}
    \vspace{-7mm}
   \caption{Three scenes from the movie \textit{The Godfather}. In the first scene, The Godfather (Don Vito Corleone) discusses family business affairs with his eldest son Sony and family lawyer Tom Hagen. Then the movie transitions into the second scene, where the mafia family meets Virgil Sollozzo to hear his business proposition. Afterward, in the third scene, Don Vito chides his son Sonny for interfering in the middle of his talk. Generally, movies are composed of well-defined scenes, and detecting those scenes is essential to high-level movie understanding. Therefore, this work aims to detect high-level scenes from a long-range movie video.}
   \vspace{-4mm}
   \label{fig:intro}
\end{figure*}

Imagine watching \textit{The Godfather} movie as illustrated in Figure~\ref{fig:intro}.  The first scene starts in the office room of mafia boss Don Vito Corleone, where he is discussing family business affairs with his eldest son Sonny Corleone and the family lawyer Tom Hagen. Then in the second scene, the family meets Virgil Sollozzo to hear about his business proposition. Afterward, the movie transitions into a third scene, where Don Vito chides his son Sonny for interfering. All of these movie scenes are inherently connected and create a complex storyline. Thus, the ability to recognize such movie scenes is essential for understanding the plot of a movie. Moreover, identifying movie scenes enables broader applications, such as content-driven video search, preview generation, and minimally disruptive ad insertion.

Unlike standard video recognition tasks (e.g., action recognition), which require a short-range analysis of video clips lasting only 5–10 seconds, movie scene detection requires short- and long-range understanding, where videos might last several minutes. For instance, to successfully recognize the scenes in Figure~\ref{fig:intro}, we need to identify the visual concepts in each frame, such as characters, objects, and backgrounds (short-range modeling), while also reasoning about complex temporal events and activities (long-range modeling). For example, if we only look at the boundary frames of the first and second scenes of Figure~\ref{fig:intro} and identify the local properties (e.g., characters, objects, and backgrounds), it is quite difficult to determine the transition between the two scenes. However, considering longer temporal windows with frames from both scenes, we can recognize global events and identify their boundaries. 

We also note that models that only use short-range temporal context typically produce many false positives due to abrupt low-level changes in the scene dynamics (e.g., camera switching between characters). For example, if we look at the middle two frames of scene 2 in Figure~\ref{fig:intro}, we might mistake them for a scene boundary since the frames contain two characters. However, looking at all the frames in scene 2 makes it clear that they belong to the same scene. Therefore, a successful movie scene detection model should identify the short-range visual elements and also reason about their long-range temporal dependencies in a movie segment.

However, most existing movie scene detection methods~\cite{mun2022boundary, chen2021shot} are built using convolutional neural networks (CNNs), which are inherently designed using local operators (i.e., convolution kernels) for short-range video analysis. Thus, such models often fail to capture the long-range dependencies crucial for accurate movie scene detection. Recently, transformers have been shown to effectively model short-range and long-range dependencies in various Natural Language Processing (NLP) tasks~\cite{bahdanau2014neural, vaswani2017attention, devlin2018bert, liu2019roberta, yang2019xlnet, dai2019transformer, brown2020language, raffel2019exploring}. Inspired by the success of transformers in NLP, subsequent works have shown the remarkable success of using transformers in several short-range vision tasks~\cite{dosovitskiy2020image, carion2020end}. However, applying Transformers to long video sequences remains challenging due to the quadratic cost of self-attention. To overcome these issues, recent work proposed a structured state-space sequence (S4) operator \cite{gu2021combining, gu2021efficiently, mehta2022long} for efficient long-range sequence analysis.

Motivated by the developments in transformer and S4-based architectures, we present TranS4mer, an efficient end-to-end model for movie scene boundary detection, which combines the strengths of transformers for short-range modeling and S4 operators for long-range modeling. Specifically, we propose a novel state-space self-attention (S4A) block and build our TranS4mer model by stacking multiple S4A blocks on top of each other. Our model takes a sequence of frames divided into movie shots as input, where a shot is defined as a series of frames captured by the same camera over an uninterrupted period~\cite{thirard1994robert}. The S4A block first captures the short-range intra-shot dependencies by applying self-attention over the frames in each shot. Subsequently, it aggregates the inter-shot interactions over long movie segments by employing the state-space operator over all shots in a given movie video. 

One key property of our model is that it independently applies self-attention to each shot, significantly reducing the cost of standard self-attention. For instance, if we have a video of 25 shots and each shot contains 3 frames, applying a standard vision transformer~\cite{dosovitskiy2020image} with a patch size of $32\times32$ will result in 3,675 tokens and a single self-attention operation will require $\sim$13.5M pairwise comparisons. In contrast, applying self-attention within each shot parallelly reduces the number of comparisons  $25\times$ times. Simultaneously, TranS4mer attains long-range modeling ability through the efficient state-space layers that do not require costly pairwise comparisons and operate over the entire input video. As a result, the proposed TranS4mer model can be efficiently applied to long movie videos.

We experiment with three movie scene detection datasets: MovieNet~\cite{huang2020movienet}, BBC~\cite{baraldi2015deep}, and OVSD~\cite{rotman2016robust}, and report that our method outperforms prior approaches by large margins ($\bf +3.38\%$ AP, $\bf +4.66\%$ and $\bf +7.36\%$ respectively). TranS4mer is also $\bf 2\times$ faster and requires $\bf 3\times$ less GPU memory than the pure Transformer-based models. Moreover, to evaluate the generalizability of our model, we experiment with several long-range video understanding tasks, such as movie clip classification and procedural activity recognition. TranS4mer performs the best in 5 out of 7 movie clip classification tasks on LVU~\cite{wu2021towards}, the best in procedural activity classification on Breakfast~\cite{kuehne2014language}, and the second-best in procedural activity classification on COIN~\cite{tang2019coin} datasets.

\section{Related works}
\label{sec:related}

 \textbf{Long Sequence Modeling.} Many modern approaches to long sequence modeling started in the field of Natural Language Processing (NLP). Bahdanau \textit{et al.}~\cite{bahdanau2014neural} introduced an RNN architecture with an attention mechanism for machine translation. Afterward, Vaswani \textit{et al.}~\cite{vaswani2017attention} proposed the transformer model for capturing long-range dependencies in text sequences. Following this work, various improved transformer architectures were proposed \cite{devlin2018bert, liu2019roberta, yang2019xlnet, dai2019transformer, brown2020language, raffel2019exploring}. However, despite these advances, applying transformers to very long sequences is still challenging because of the quadratic cost of the self-attention operation. Consequently, many recent methods have developed efficient variations of standard self-attention~\cite{kitaev2020reformer, zaheer2020big, katharopoulos2020transformers, choromanski2020rethinking, patrick2021keeping}. Furthermore, the recent work of Gu \textit{et al.}~\cite{gu2021combining, gu2021efficiently, goel2022sashimi} proposed a structured state-space sequence (S4) model for long sequence analysis. The S4 models have achieved impressive results in many NLP tasks \cite{gu2021combining, gu2021efficiently, goel2022sashimi, mehta2022long, gupta2022diagonal}, while having only linear memory and computation costs (w.r.t the sequence length). Building on this work, Islam \textit{et al.}~\cite{mohaiminul2022long} proposed a model that applies S4 on pre-extracted video features for long-range video classification. In contrast, we propose an end-to-end trainable movie scene detection model that combines S4 and self-attention into a novel S4A block, which we use to build our TranS4mer model. %for the challenging movie scene detection task. 

 \textbf{Movie Scene Detection.} The early approaches for movie scene detection consisted predominantly of unsupervised clustering-based methods \cite{rotman2016robust, zhou2003constructing, baraldi2015shot, chasanis2008scene, rasheed2005detection, sidiropoulos2011temporal, baraldi2015analysis} built using hand-crafted features. Recently, Huang \textit{et al.}~\cite{huang2020movienet} introduced a large-scale dataset for movie scene detection called MovieNet, which became the most popular benchmark for this problem. Subsequently, Huang \textit{et al.}~\cite{huang2020movienet} and Rao \textit{et al.}~\cite{rao2020local} proposed an LSTM-based model for movie scene detection, which is applied on top of pre-extracted CNN features. Most recently, Chen \textit{et al.}~\cite{chen2021shot} and Mun \textit{et al.}~\cite{mun2022boundary} introduced several end-to-end movie scene detection approaches built using 2D/3D CNN architectures. However, these CNN-based models are limited by the smaller receptive fields of the local convolution operations. In contrast, our TranS4mer model is constructed using global operators (state-space and self-attention), which are more effective for long-range modeling.

\begin{figure}[t]
    \vspace{-3mm}
  \centering
    \includegraphics[width=1.0\linewidth]{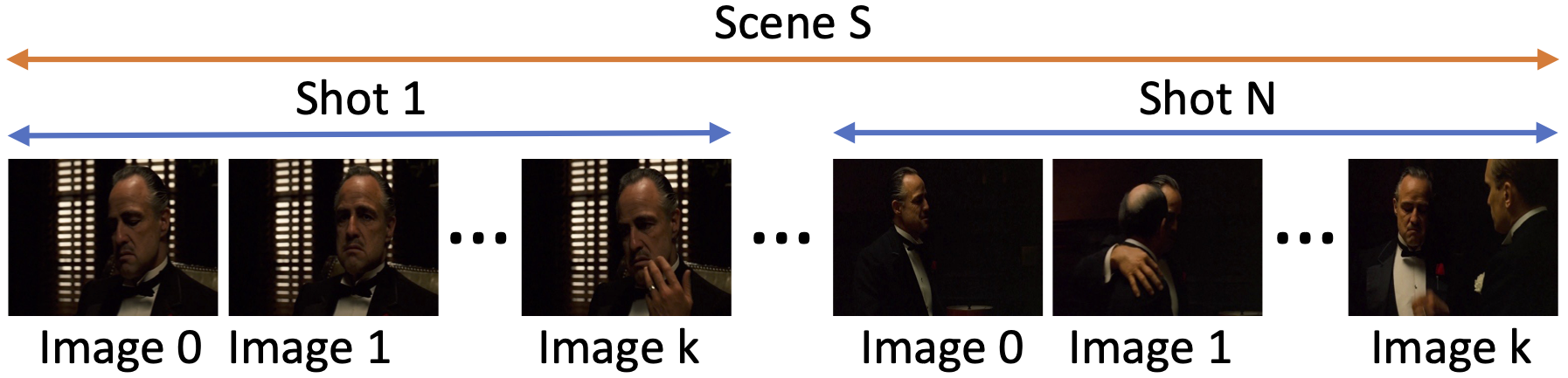}
    \vspace{-6mm}
   \caption{The hierarchical structure of long movie videos. First, multiple consecutive frames captured from the same camera position depict a shot. Next, multiple neighboring shots constitute a movie scene. Finally, a movie is composed of a collection of scenes. Here, we show one scene that contains N shots with $K$ frames in each shot.}
   \vspace{-5mm}
   \label{fig:shot}
\end{figure}

\section{Background}
\label{sec:background}

 \textbf{The State-Space Model} (SSM) is a fundamental model used in many scientific fields, such as control theory and computational neuroscience. We can define the SSM in continuous time using the following equations:
 
\begin{equation}
  \label{eq:ssm}
  \begin{aligned}
    h'(t) &= A h(t) + B x(t) \\
    y(t)  &= C h(t) + D x(t)
  \end{aligned}
\end{equation}

Here, $x(t)$ is the input signal, $h(t)$ is a latent state, $h'(t)$ is the first order derivative of $h(t)$ wrt $t$, i.e., $h'(t)=\frac{dh}{dt}$, $y(t)$ is the output signal, and $A$ is the state matrix. Equation~\eqref{eq:ssm} can be discretized by sampling the underlying input signal $u(t)$ with a step size $\Delta$.
\begin{equation}
    \label{eq:ssm-recurrence}
    \begin{aligned}
    h_k &= \overline{A} h_{k-1} + \overline{B} x_k \\
    y_k  &= \overline{C} h_k + \overline{D} x_k \\
    %\overline{A} &= (I - \Delta/2 \cdot A)^{-1}(I + \Delta/2 \cdot A)
    \end{aligned}
\end{equation}

Here, $\overline{A}$, $\overline{B}$, $\overline{C}$, and $\overline{D}$ are the discrete versions of $A,B,C$ and $D$. Furthermore, the equation~\eqref{eq:ssm-recurrence} can be expressed as a convolution operation ($*$):
\begin{equation}
  \label{eq:ssm-convolution}
\begin{aligned}
   y = \overline{K} \ast x \qquad
  \overline{K} =
  (\overline{C}\overline{B}, \overline{C}\overline{A}\overline{B}, \dots, \overline{C}\overline{A}^{L-1}\overline{B}),
\end{aligned}
\end{equation}
where $\overline{K}$ is a kernel, and $L$ is the sequence length.

 \textbf{S4 Model.} The Structured State Space (S4) model is a particular class of SSM that requires the state matrix $A$ to be \textit{diagonal and low-rank}. Because of the specific structure of the state matrix $A$, we can derive a closed-form formula for the convolutional kernel $\overline{K}$ in eqaution~\eqref{eq:ssm-convolution}~\cite{gu2021efficiently}, which then facilitates an efficient computation of $\overline{K}$ without having to multiply the matrix $A$, $L-1$ times.

 \textbf{Gated S4 Model.} Borrowing the idea from Gated Attention~\cite{hua2022transformer}, Metha \textit{at el.}~\cite{mehta2022long} introduced a Gated S4 model, which first uses linear layers to compute two latent representations ($u$ and $v$) from the input sequences $x$, and then applies an S4 layer to obtain a contextualized representation $\hat{u}$. Finally, the tokens from $\hat{u}$ are used to gate $v$:
\begin{equation}
  \label{eq:gs4}
\begin{aligned}
    u &= \phi(W_ux) \qquad
    v = \phi(W_vx) \\
    h &= S4(u)  \qquad\quad
    \hat{u} = W_hx \\
    y &= W_y(\hat{u} \odot v)
\end{aligned}
\end{equation}
Here, $W_u, W_v, W_h, W_y$ are the weight matrices of linear layers, $\phi$ denotes GELU nonlinearity~\cite{hendrycks2016bridging}, $\odot$ is elementwise multiplication, and S4 denotes an S4 layer described above. %It is worth noting that the dimension of $u$ is much smaller than the dimension of the input $x$, which further reduces the computation cost of the S4 operator.

\section{Technical Approach}
\label{sec:methodology}

\begin{figure*}[t]
  \centering
   \vspace{-5mm}
    \includegraphics[width=0.95\linewidth]{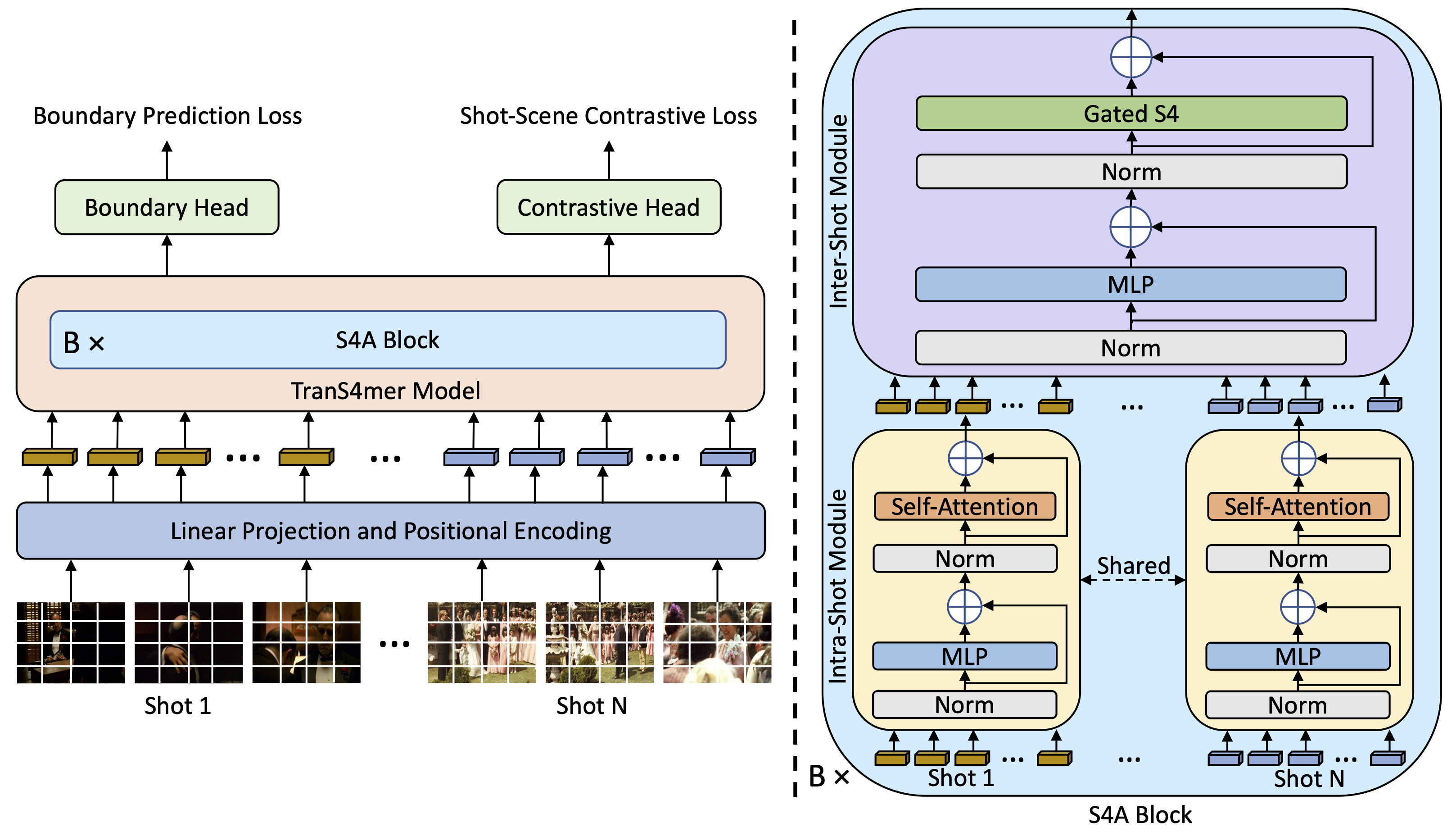}
    \vspace{-3mm}
   \caption{Overview of the TranS4mer model. \textbf{Left:} Our model takes a sequence of frames divided into automatically-detected shots. Following the ViT~\cite{dosovitskiy2020image}, the frames are decomposed into non-overlapping patches and passed through a linear projection layer. Afterward, we feed these patch-level embeddings through our TranS4mer model, which consists of $B$ state-space self-attention (S4A) blocks stacked on top of each other. Finally, the CLS tokens of all shots are passed to two linear layers, i.e., the boundary prediction head (which outputs a binary label indicating the boundary shot) and the contrastive head (which maximizes intra-shot similarity). \textbf{Right:} Each of our proposed S4A blocks contains a self-attention-based intra-shot module that models the interaction among frames in the same shot and a Gated S4-based inter-shot module that aggregates long-range temporal cues from shots far away from each other.}
   \vspace{-4mm}
   \label{fig:method}
\end{figure*}

We aim to develop an efficient scene detection method for long movie videos. Movie videos have a specific hierarchical structure consisting of three semantic levels: (i) frames, (ii) shots, and (iii) scenes (See Figure~\ref{fig:shot}). First, multiple consecutive frames constitute a shot, where the frames are captured from the same camera position over an uninterrupted period~\cite{thirard1994robert}. Second, multiple consecutive shots construct a scene, where a scene is a semantically cohesive unit of the storyline~\cite{rui1999constructing, filmencyclopedia}. Finally, a movie is composed of a collection of scenes. Shots can be easily recognized by detecting the changes in low-level visual cues~\cite{sidiropoulos2011temporal, cotsaces2006video, castellano2018pyscenedetect}. However, identifying the scene boundary is more challenging as it requires high-level semantic understanding.

 \textbf{Problem Overview.} First, we formally define the problem of movie scene detection. We are given a movie video $V$ divided into $T$ non-overlapping automatically-detected shots, $V = \{s_1, \dots, s_i, \dots, s_T\}$. We can then uniformly sample $K$ frames from each shot. Therefore, a shot is represented as $s_i=\{f_1, \dots, f_j, \dots, f_K\}$, where $f_j\in \mathbb{R}^{3\times H\times W}$ depicts an RGB frame, $H$ is the height, and $W$ is the width of each frame. Following these definitions, we can then formulate the movie scene detection problem as a prediction task, where we need to produce a binary output $y_i\in \{0, 1\}$ for each video shot $s_i$ (i.e., whether a given shot is at a movie scene boundary or not). 

 \textbf{Approach Overview.} Our movie scene detection model, called the State-Space Transformer (TranS4mer), is built using our proposed S4A blocks that combine the strengths of structured state-space sequence (S4) and self-attention (A) layers. The S4A block utilizes the self-attention layer to encode short-range dependencies between frames in each shot and the S4 layer for capturing long-range dependencies between shots far away from each other. A detailed illustration of our model is presented in Figure~\ref{fig:method}. Below, we describe each component of our model in more detail.

\subsection{The TranS4mer Model}

Following previous methods\cite{rao2020local, chen2021shot, mun2022boundary}, our model takes as input $N=2m+1$ temporally adjacent shots, $\mathcal{V}_i=\{s_{i-m}, \dots, s_i, \dots, s_{i+m}\}$, and outputs a binary prediction $y_i\in \{0, 1\}$ for the shot $s_i$. The neighboring shots provide long-range contextual cues for predicting the boundary at shot $s_i$. Here, $\mathcal{V}_i\in{\mathbb{R}^{N\times K\times 3\times H\times W}}$ is composed of $K$ frames from each of $N$ consecutive shots, where $H$ is the height, and $W$ is the width of each frame. 

Following ViT\cite{dosovitskiy2020image}, we divide each frame into $P$ non-overlapping patches of size $p\times p$, where $P=HW/p^2$. Then we use a linear layer to project each patch into a hidden dimension $D$, and a positional embedding is added to each patch token. As a result, we obtain a tensor of shape $\mathbf{V}_i\in\mathbb{R}^{N\times K\times P\times D}$, which is fed to our TranS4mer model. 

 \textbf{The S4A Block.} Our proposed S4A block enables us to compute a strong short-range feature representation and model long-range temporal dependencies among shots. It consists of two modules: a short-range intra-shot module and a long-range inter-shot module.

 \textit{Intra-Shot Module.} The intra-shot module takes the input tensor $\mathbf{V}_i\in\mathbb{R}^{N\times K\times P\times D}$, which we can think of as a collection of $N$ shots, where a sequence of tokens represents each shot $S_i={x_1, \dots, x_L}$, where $x_i\in \mathbb{R}^D$, $L = K\times P$. The intra-shot module models the sequence of tokens from each shot independently. To this end, it first applies a multi-layer perceptron (MLP) layer and a residual connection. Afterward, it applies a Multiheaded Self Attention (MSA) layer and another residual connection. Layer Normalization (LN) is applied before the MLP and the MSA layers. The module can be defined as follows:
 
\vspace{-2mm}
\begin{equation}
\begin{aligned}
    &\mathbf{x'} = \text{MHA(LN(}\mathbf{x}_{in}\text{))} + \mathbf{x}_{in}\\
    &\mathbf{x}_{out} = \text{MLP(LN(}\mathbf{x}^\prime\text{))} + \mathbf{x'}\\
\end{aligned}
\end{equation}
\vspace{-2mm}

The output of the intra-shot module $\mathbf{V'}_i\in\mathbb{R}^{N\times K\times P\times D}$, is then passed to the inter-shot module.

%Note that the self-attention operation has a quadratic cost, making it challenging to apply it to long shot sequences. For instance, applying self-attention to the entire sequence of tokens in $\mathbf{V_i}$ incurs a cost of $O(N^2L^2)$. Instead, applying self-attention to tokens from each shot separately reduces the computation cost to $O(NL^2)$. 

Note that applying self-attention to tokens from each shot separately reduces the computation cost to $O(NL^2)$. In contrast, applying self-attention to the entire sequence of tokens in $\mathbf{V_i}$ would cost $O(N^2L^2)$.

 \textit{Inter-Shot Module.} Our previously defined intra-shot module only models short-range dependencies and, thus, lacks the ability for long-range temporal understanding, which is crucial for the movie scene detection task. To address this limitation, we propose an efficient inter-shot module based on the Gated S4 operator that aggregates long-range inter-shot cues. 

First, we unroll the input tensor $\mathbf{V'}_i$ into a sequence ${z_1, \dots, z_{L'}}$, where $z_i\in \mathbb{R}^D$, $L' = N\times K\times P$. Then, we pass the sequence to a multi-layer perceptron (MLP) layer and a residual connection. Afterward, we apply a Gated S4 (GS4) layer and another residual connection. Layer Normalization (LN) is applied before MLP and GS4 layers:

\vspace{-2mm}
\begin{equation}
\begin{aligned}
    &\mathbf{z'} = \text{GS4(LN(}\mathbf{z}_{in}\text{))} + \mathbf{z}_{in}\\
    &\mathbf{z}_{out} = \text{MLP(LN(}\mathbf{z}^\prime\text{))} + \mathbf{z'}\\
\end{aligned}
\end{equation}
%\vspace{-3mm}
\vspace{-0.1cm}

The output of this operation is then passed into the next S4A block. Unlike the intra-shot module, which operates on a short sequence of length $L=K\times P$, the inter-shot module operates on much longer sequences of $L'=N\times K\times P$. Our model can efficiently process such long sequences due to the Gated S4 layers in the inter-shot module.

%Our model can efficiently process such long sequences because the inter-shot module uses the Gated S4 layer with a linear computation cost w.r.t. sequence length.

\subsection{Training and Loss Functions} 

Most prior works in move scene detection use self-supervised pretraining~\cite{mun2022boundary, chen2021shot}. Thus, we also adopt the self-supervised pretraining strategy of Mun \textit{at el.}~\cite{mun2022boundary}, which uses pseudo boundaries of move scenes from the Dynamic Time Warping algorithm as a supervisory signal. Afterward, we finetune our pretrained model using the labeled subset of the data.  Note that we do not claim technical novelty over the pretraining strategy. Instead, our main goal is to develop an efficient state-space transformer backbone for movie scene detection. Below, we provide more details related to the pretraining and fine-tuning protocols.

%Therefore, we adopt the most recent pretraining/finetuning protocols~\cite{mun2022boundary} for evaluating our proposed architecture. Below, we provide more details about the pretraining and finetuning techniques.

 \textbf{Pretraining.}  Leveraging the pseudo movie scene boundaries, we optimize our model using the following two losses: (i) a shot-scene contrastive loss and (ii) a pseudo-boundary prediction loss. The contrastive loss maximizes the similarity between the representations of a shot and its corresponding pseudo ground truth scene while also minimizing the similarity of a shot with all the other scenes. We represent each shot with its respective CLS token, whereas each scene is represented by mean-pooling the CLS tokens of all shots in that scene. The DTW algorithm divides the input video $\mathcal{V}_i=\{s_{i-m}, \dots, s_i, \dots, s_{i+m}\}$ into two pseudo-scenes $\mathcal{V}_L=\{s_{i-m}, \dots, s_{i^*}\}$ and $\mathcal{V}_R=\ s_{i^*+1}, \dots, s_{i+m}\}$, here $s_{i^*}$ is the pseudo-boundary shot. We can then define the shot-scene contrastive loss as:%\GB{This equation does not look great

 \vspace{-0.3cm}
\begin{equation}
\begin{aligned}
&l_\text{C}(\mathbf{r}, \bar{\mathbf{r}})=-\log \frac{\mathcal{S}(\mathbf{r}, \bar{\mathbf{r}})}{\mathcal{S}( \mathbf{r}, \bar{\mathbf{r}}) + \sum\limits_{\mathbf{r_n}} \mathcal{S}(\mathbf{r}_n, \bar{\mathbf{r}}) + \sum\limits_{\bar{\mathbf{r}}_n}\mathcal{S}(\mathbf{r}, \bar{\mathbf{r}}_n)}\\
%&\mathcal{L}_{\text{C}} = l_\text{C} (\mathbf{r}_{i-m}, \bar{\mathbf{r}}_{L})+l_ \text{C} ( \mathbf{r}_{i+m},\bar{\mathbf{r}}_{R})\\
\end{aligned}
\end{equation}
\vspace{-0.3cm}

Here, $\mathbf{r}$ is a shot representation obtained by applying a linear layer on the CLS token of that shot, $\bar{\mathbf{r}}_{L}$ and $\bar{\mathbf{r}}_{R}$ are the scene-level representations of pseudo-scenes $\mathcal{V}_L$ and $\mathcal{V}_R$, $\mathbf{r}_n$ and $\bar{\mathbf{r}}_n$ are the negative shots and scenes from the mini-batch. Similarity between two representations is calculated as: $\mathcal{S}(\mathbf{x}, \mathbf{y}) = \exp((\mathbf{x}^\top \mathbf{y})/(\lVert \mathbf{x}\rVert \lVert \mathbf{y}\rVert))$. The first loss term can then be written as: $\mathcal{L}_{\text{C}} = l_\text{C} (\mathbf{r}_{i-m}, \bar{\mathbf{r}}_{L})+l_ \text{C} ( \mathbf{r}_{i+m},\bar{\mathbf{r}}_{R})$.

The second loss function of pseudo-boundary prediction is a simple objective that guides the model to predict a pseudo-boundary from a given shot representation. We can define this loss as a binary cross-entropy loss:

\vspace{-0.2cm}
\begin{equation}
    \mathcal{L}_{\text{B}} = -\log \left( \rho_b(\mathbf{r}_{i^{*}}) \right) - \log \left( 1 - \rho_b(\mathbf{r}_{\bar{b}}) \right)
\end{equation}
\vspace{-0.2cm}

Here, $\mathbf{r}_{i^{*}}$ is the CLS representation of the pseudo-boundary shot $s_{i^*}$, $\mathbf{r}_{\bar{b}}$ is the representation of a randomly selected non-boundary shot $s_{\bar{b}}$, $\rho_b$ is a projection layer that outputs the probability of a shot being the pseudo-boundary.

The overall pretraining loss is the sum of the shot-scene contrastive loss and the pseudo-boundary loss.

 \textbf{Finetuning.} After the self-supervised pretraining, we finetune our model using the subset of the dataset with ground truth annotations. We use a binary cross-entropy loss as in~\cite{mun2022boundary} to finetune our model.

\subsection{Implementation Details}

Following the ViT~\cite{dosovitskiy2020image}, we resize video frames to a spatial dimension of $224\times 224$ and divide using a patch size of $32\times 32$. Our model contains 12 S4A blocks and uses a hidden dimension of 384. It takes 25 neighboring shots as input, where each shot contains 3 frames. We use four data augmentation techniques: random crop, flip, color jitter, and Gaussian blur. We use a batch size of 256 and pretrain our model for 10 epochs before finetuning for 20 epochs. During pretraining, we use a base learning rate of 0.3, which we reduce using a cosine schedule after a linear warmup strategy of 1 epoch. During finetuning, we use a learning rate of $10^{-6}$. We use the Adam~\cite{kingma2014adam} optimizer with a momentum of 0.9 and weight decay of $10^{-6}$. We train our models using 8 NVIDIA RTX A6000 GPUs.

%\GB{Do these provide a description for our S/32 default model? If not it should}

\begin{table*}[t]
    %\vspace{-7mm}
  \centering
  \begin{tabular}{@{}llcccccc@{}}
    &Method & AP ($\uparrow$) & mIoU ($\uparrow$) & AUC-ROC ($\uparrow$) & F1 ($\uparrow$) & Memory (GB) ($\downarrow$) & Samples/s ($\uparrow$)\\
    \toprule
    \multirow{10}{*}{Unsupervised}&
    GraphCut~\cite{rasheed2005detection} & 14.10 & 29.70 & - & - & - & -  \\&
    SCSA~\cite{chasanis2008scene} & 14.70 & 30.50 & - & - & - & - \\&
    DP~\cite{han2011video} & 15.50 & 32.00 & - & - & - & - \\&
    Story Graph~\cite{tapaswi2014storygraphs} & 25.10 & 35.70 & - & - & - & - \\&
    Grouping~\cite{rotman2016robust} & 33.60 & 37.20 & - & - & - & - \\&
    BaSSL~\cite{mun2022boundary} & 31.55 & 39.36 & 71.67 & 32.55 & 34.28 & 0.96 \\\cline{2-8}&
    Transformer~\cite{dosovitskiy2020image} & 32.16 & 37.24 & 70.23 & 31.24 & 30.28 & 1.27\\&
    TimeSformer~\cite{gberta_2021_ICML} & 32.47 & 37.76 & 70.95 & 31.45 & 28.12 & 1.47 \\&
    Vanilla S4~\cite{gu2021efficiently} & 33.34 & 38.12 & 71.86 & 32.21 & 15.62 & 1.83 \\&
    TranS4mer & \textbf{34.45} & \textbf{39.60} & \textbf{73.25} & \textbf{33.41} & \textbf{10.13} & \textbf{2.57}\\
    \midrule
    \multirow{10}{*}{Supervised}&
    Siamese~\cite{baraldi2015deep} & 35.80 & 39.60 & - & - & - & - \\&
    MS-LSTM~\cite{huang2020movienet} & 46.50 & 46.20 & - & - & - & - \\&
    LGSS~\cite{rao2020local} & 47.10 & 48.80 & - & - & - & - \\&
    ViS4mer~\cite{mohaiminul2022long} & 55.13 & 48.27 & 88.74 & 46.15 \\&
    ShotCoL~\cite{chen2021shot} & 53.40 & - & - & - & 34.28 & 0.96 \\&
    BaSSL~\cite{mun2022boundary} & 57.40 & 50.69 & 90.54 & 47.02 & 34.28 & 0.96 \\\cline{2-8}&
    %BaSSL\dag & 56.75 & 49.57 & 90.10 & 45.74 \\
    Transformer~\cite{dosovitskiy2020image} & 58.81 & 51.21 & 90.84 & 47.88 & 30.28 & 1.27\\&
    TimeSformer~\cite{gberta_2021_ICML} & 59.62 & 50.75 & 90.66 & 48.02 & 28.12 & 1.47 \\&
    Vanilla S4~\cite{gu2021efficiently} & 59.71 & 51.32 & 90.96 & 47.85 & 15.62 & 1.83 \\&
    TranS4mer & \textbf{60.78} & \textbf{51.91} & \textbf{91.89} & \textbf{48.36} & \textbf{10.13} & \textbf{2.57}\\
    \bottomrule
    %\dag our implementation
  \end{tabular}
  \vspace{-0.1cm}
  \caption{\textbf{Results on MovieNet~\cite{huang2020movienet}.} Horizontal lines in both Unsupervised and Supervised rows separate prior works from our implemented baselines. First, our method achieves state-of-the-art performance, outperforming the previous best method (BaSSL) by a significant margin of \textbf{3.38}\% AP. Second, compared to prior unsupervised approaches, the unsupervised TranS4mer model achieves better performance in all metrics. Finally, TranS4mer is $\textbf{2.68}\times$ faster and requires $\textbf{3.38}\times$ less GPU memory than the previous SOTA methods (ShotCol and BaSSL), while also being faster and more memory efficient than the transformer and state-space baselines.
  \vspace{-0.2cm}
  }
  \label{tab:main result}
\end{table*}

\section{Experimental Setup} 
\label{sec:Experimental setup}

\subsection{Datasets} 
\label{sec:datasets}

 \textbf{MovieNet}~\cite{huang2020movienet} is a large-scale dataset containing 1100 movies with 1.6 million shots, and 318 of these movies have ground-truth scene boundary annotation. The annotated subset of the dataset is divided into 190, 64, and 64 movies for training, validation, and testing. Following the previous approaches, we pretrain our model using all the movies, finetune using the standard training split of the annotated subset, and report the result on the test split.

 \textbf{BBC}~\cite{baraldi2015deep} contains 11 episodes from the BBC TV series \textit{Planet Earth}~\cite{planetearth}. The average duration of the videos is 50 minutes. The dataset contains 670 scenes and 4.9K shots.

 \textbf{OVSD}~\cite{rotman2016robust} contains 21 short films with an average duration of $~30$ minutes. It has a total of 10K shots and 300 scenes extracted from movie scripts.

\subsection{Evaluation Metrics}

 \textbf{AP} measures the weighted average of precisions at different thresholds.

 \textbf{mIoU} is the average intersection over union of predicted scene boundaries with the closest ground truth boundaries.

 \textbf{AUC-ROC} is calculated as the True Positive Rate with respect to the False Positive Rate at different thresholds. 

 \textbf{F1-Score} is the harmonic mean of precision and recall.

\subsection{Baselines}
\label{sec:baselines}
\vspace{-0.1cm}

 \textbf{Prior Methods.} In our comparisons, we include several unsupervised approaches such as GraphCut~\cite{rasheed2005detection}, SCSA~\cite{chasanis2008scene}, DP~\cite{han2011video}, Story Graph~\cite{tapaswi2014storygraphs}, and Grouping~\cite{rotman2016robust}. These methods use various clustering techniques using pre-extracted features. Moreover, we compare our approach with supervised approaches, such as Siamese~\cite{baraldi2015deep}, MS-LSTM~\cite{huang2020movienet}, and LGSS~\cite{rao2020local}, which also use pre-extracted CNN features. We also compare against the long-range video model ViS4mer~\cite{mohaiminul2022long}, which uses S4 layers as a late-fusion strategy. Finally, we include two recent end-to-end trainable models that use self-supervised pretraining, namely, ShotCoL~\cite{chen2021shot} and BaSSL~\cite{mun2022boundary}. Both models use CNN models based on ResNet-50~\cite{he2016deep}. 

%\vspace{-0.2cm}
 \textbf{Our Implemented Baselines.} To make our comparisons more complete, we implement three additional baselines: \underline{\textit{1) Transformer:}} We construct this baseline by following the standard Vision Transformer~\cite{dosovitskiy2020image}. In particular, we replace the Gated S4 layers of the TranS4mer model with self-attention layers and keep everything else unchanged. \underline{\textit{2) TimeSformer:}} We construct this baseline by replacing the Gated S4 layers of our model with the divided space-time attention proposed by Bertasius \textit{at el.}~\cite{gberta_2021_ICML}. \underline{\textit{3) Vanilla S4:}} We develop this baseline by replacing the Gated S4 layers of the TranS4mer model with the S4 layers ~\cite{gu2021efficiently}.

\section{Results and Analysis}
\label{sec:results}

\subsection{Main Results on the MovieNet Dataset} 

 \textbf{Comparison to the State-of-the-Art.} We present the results on  MovieNet~\cite{huang2020movienet} in Table~\ref{tab:main result}. We first compare our method with the unsupervised approaches, including GruphCut~\cite{rasheed2005detection}, SCSA~\cite{chasanis2008scene}, DP~\cite{han2011video}, Story Graph~\cite{tapaswi2014storygraphs}, Grouping~\cite{rotman2016robust}, and BaSSL (unsupervised)~\cite{mun2022boundary}. For a fair comparison, we train an unsupervised variant of our TranS4mer model using pseudo-boundary loss without ground-truth annotations and report the result without supervised finetuning. Our results suggest that our unsupervised TranS4mer model achieves the best performance according to all metrics and outperforms the prior best method Grouping by \textbf{+0.85\%} AP. Next, we compare the supervised variant of our model (i.e., finetuned with ground truth annotations) with the state-of-the-art supervised methods (ShotCol~\cite{chen2021shot} and BaSSL~\cite{mun2022boundary})). We also compare with the state-space-based long-range video classification model ViS4mer~\cite{mohaiminul2022long}. We observe that TranS4mer achieves the best performance in all metrics and outperforms the previous best method (BaSSL) by a large margin (\textbf{+3.38}\% AP).

 \textbf{Comparison with Transformer and State-space Baselines.} In Table~\ref{tab:main result}, we also compare our TranS4mer model with the Transformer and State-space baselines: Transformer, TimeSformer, and Vanilla S4 (See subection~\ref{sec:baselines}). Note that all three baselines use intra-shot and inter-shot modules; however, instead of a Gated S4 layer, Transformer uses standard self-attention~\cite{dosovitskiy2020image}, TimeSformer uses divided space-time attention~\cite{gu2021efficiently}, and Vanilla S4 uses standard S4~\cite{gu2021efficiently} at the inter-shot module. Moreover, similar to the TranS4mer-Unsupervised method, we train unsupervised variants of these baselines without any ground-truth annotation. First, we observe that all three baselines outperform the previous state-of-the-art method (BaSSL~\cite{mun2022boundary}) in both supervised and unsupervised settings, indicating the effectiveness of our proposed intra-shot and inter-shot block designs. Second, TranS4mer performs better than all three of these baselines, outperforming Transformer by \textbf{+1.97}\% and \textbf{+2.29}\%, TimeSformer by \textbf{+1.16}\% and \textbf{+1.98}\%, and Vanilla S4 by \textbf{+1.07}\% and \textbf{+1.11}\% in supervised and unsupervised approaches, respectively. These experiments indicate that TranS4mer is better suited for movie scene detection than the standard self-attention and state-space models.
 
 \textbf{Computational Cost Analysis.} We compare the GPU memory consumption (in gigabytes) and the training speed (samples/s) of our model with the previous best methods as well as our implemented baselines in Table~\ref{tab:main result}. First, we observe that, in addition to achieving the best performance, our method also requires $\mathbf{3.38\times}$ less memory and is {$\mathbf{2.68\times}$} faster than the previous state-of-the-art method BaSSL~\cite{mun2022boundary}. Second, compared to the standard Transformer baseline, TranS4mer is $\mathbf{2\times}$ faster and requires $\mathbf{3\times}$ less GPU memory. Finally, our model is significantly cheaper than efficient self-attention and state-space models such as TimeSformer and Vanilla S4. Whereas TimeSformer and Vanilla S4 require \textbf{28.12} GB and \textbf{15.62} GB of memory, our TranS4mer requires only \textbf{10.13} GB of memory. Moreover, the training speed of TranS4mer is \textbf{2.57} samples/second compared to the \textbf{1.47} and \textbf{1.82} samples/second speed of TimeSformer and Vanilla S4, respectively. These observations suggest that  TranS4mer is much more efficient than prior methods and standard self-attention and state-space-based models.

 \textbf{Comparison Using the Same Backbone Model.} For a fair comparison, we also experiment with a previous-best BaSSL~\cite{mun2022boundary} method that uses the same backbone as our model (i.e., ViT-S/32). We report that our TranS4mer outperforms this BaSSL variant by \textbf{+2.77}\% AP. 
 
\subsection{Scene Detection on Other Datasets}

We experiment with two additional movie scene detection datasets, BBC\cite{baraldi2015deep} and OVSD\cite{rotman2016robust}. We present these results in Table~\ref{tab:other_datasets}, which shows that TranS4mer achieves the best result among all methods, outperforming the previous best BaSSL~\cite{mun2022boundary} method by $\bf 4.66\%$ AP on the BBC and $\bf 7.36\%$ AP on the OVSD datasets. These results demonstrate our method's ability to generalize to other datasets.

\begin{table}[t]\centering
\vspace{-3mm}
% subfloat ############
\resizebox{0.47\textwidth}{!}{%
\subfloat[\normalsize{Performance on BBC~\cite{baraldi2015deep}.}\label{tab:bbc}]{
    \begin{tabular}{l c}
        Method & AP ($\uparrow$)\\
        \toprule
        BaSSL\cite{mun2022boundary} & 39.98\\
        Transformer~\cite{dosovitskiy2020image} & 41.86\\
        TimeSformer~\cite{gberta_2021_ICML} & 42.23\\
        Vanilla S4~\cite{gu2021efficiently} & 42.56\\
        TranS4mer & \textbf{43.64} \\
        \bottomrule
    \end{tabular}
}\hspace{2mm}
\subfloat[\normalsize{Performance on OVSD~\cite{rotman2016robust}.}\label{tab:ovsd}]{
    \begin{tabular}{l c}
        Method & AP ($\uparrow$)\\
        \toprule
        BaSSL\cite{mun2022boundary} & 28.68\\
        Transformer~\cite{dosovitskiy2020image} & 33.12\\
        TimeSformer~\cite{gberta_2021_ICML} & 33.87\\
        Vanilla S4~\cite{gu2021efficiently} & 34.21\\
        TranS4mer & \textbf{36.04} \\
        \bottomrule
    \end{tabular}
}
}
%\vspace{-3mm}
\vspace{-0.2cm}
\caption{\textbf{Scene Detection on additional datasets.} TranS4mer outperforms the prior method (BaSSL) by a large margin of $\bf 4.66\%$ AP on BBC~\cite{baraldi2015deep} and $\bf 7.36\%$ AP on OVSD~\cite{rotman2016robust} datasets.
% \vspace{-1mm}
}
\label{tab:other_datasets}
\end{table}

\begin{table}
    % \vspace{-0.2cm}
    \centering
    \Huge
    % \begin{tabular}{p{1.2cm} p{0.7cm}  p{0.5cm}  p{0.5cm}  p{0.7cm}  p{0.5cm}  p{0.5cm}  p{0.5cm}}
    \resizebox{.47\textwidth}{!}{%
    \begin{tabular}{lccccccc}
     Method & Relation & Speak & Scene & Director & Genre & Writer & Year\\
    \hline
     ObjTrans.\cite{wu2021towards} & 53.10 & 39.40 & 56.90 & 51.20 & 54.60 & 34.50 & 39.10\\
     ViS4mer~\cite{mohaiminul2022long} & 57.14 & \textbf{40.79} & 67.44 & 62.61 & 54.71 & \textbf{48.80} & 44.75\\
     \textbf{TranS4mer} & \textbf{59.52} & 39.21 & \textbf{70.93} & \textbf{63.86} & \textbf{55.85} & 46.93 & \textbf{45.45}\\
     \hline
     \multicolumn{8}{c}{(a) Performance on LVU~\cite{wu2021towards}}\\
    \end{tabular}
    }
    \resizebox{.23\textwidth}{!}{%
    \begin{tabular}{lcc}
         Model & \#Data($\downarrow$) & Acc.($\uparrow$) \\
         \hline
        % VideoGraph & 306K & 69.50 \\
        % Timeception & 306K & 71.30 \\
        GHRM~\cite{zhou2021graph} & 306K & 75.50 \\
        Dist.Sup.~\cite{lin2022learning} & \bf 136M & \underline{89.90} \\
        ViS4mer\cite{mohaiminul2022long} & 495K & 88.17 \\
        \textbf{TranS4mer} & 495K & \textbf{90.27}\\
        \hline
        \multicolumn{3}{c}{(b) Performance on Breakfast~\cite{kuehne2014language}}\\
    \end{tabular}
    }
    \resizebox{.23\textwidth}{!}{%
    \begin{tabular}{lcc}
        Model & \#Data($\downarrow$) & Acc.($\uparrow$) \\
        \hline
        TSN~\cite{tang2020comprehensive} & 306K & 73.40 \\
        Dist.Sup.~\cite{lin2022learning} & \bf 136M & \textbf{90.00} \\
        ViS4mer\cite{mohaiminul2022long} & 495K & 88.41 \\
        \textbf{TranS4mer} & 495K & \underline{89.23}\\
        \hline
        \multicolumn{3}{c}{(c) Performance on COIN~\cite{tang2019coin}}\\
    \end{tabular}
    }
    \vspace{-0.3cm}
    \caption{\textbf{Long-range video classification.} (a) TranS4mer achieves best performance in 5 out of 7 classification tasks in LVU benchmark~\cite{wu2021towards}. Moreover, (b) it performs best on the Breakfast~\cite{kuehne2014language} and (c) second-best on the COIN~\cite{tang2019coin} datasets, while using significantly less pretraining data (middle column).}
    \vspace{-3mm}
\label{tab:video_classification}
\end{table}

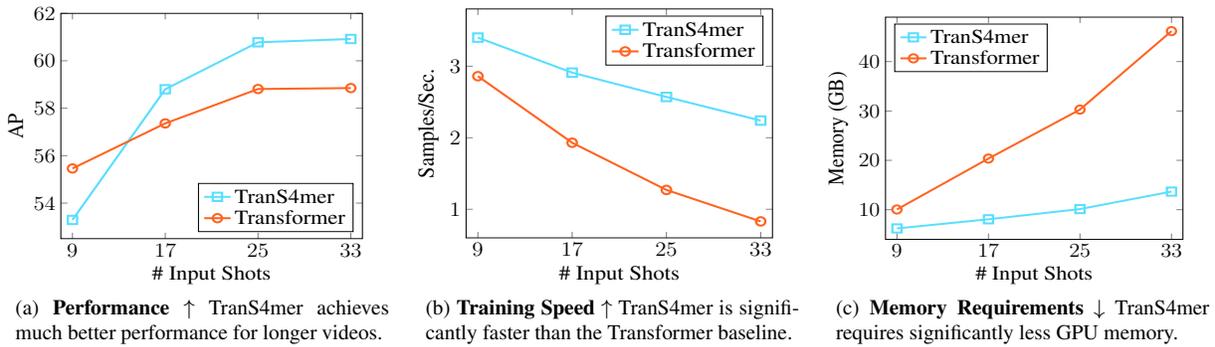
\begin{figure*}[t]\centering
\vspace{-7mm}
\captionsetup[subfigure]{margin={3mm,-2mm}}
\subfloat[\textbf{Performance $\uparrow$} TranS4mer achieves much better performance for longer videos. \label{fig:transformer_performance}]{{
    \resizebox{0.28\textwidth}{!}{%
        \begin{tikzpicture}[yscale=0.9, font={\Large}]
        \begin{axis}[
            xlabel={\# Input Shots},
            ylabel={AP},
            xmin=8, xmax=34,
            xtick={9, 17, 25, 33},
            ymin=52.5, ymax=62,
            legend cell align={left},
            legend pos=south east
        ]
        \addplot[line width=.5mm, color=CornflowerBlue, mark=square, mark size=3pt]
            coordinates {(9, 53.29)(17, 58.80)(25, 60.78)(33, 60.92)};
        
        \addplot[line width=.5mm, color=Orange,mark=o, mark size=3pt]
            coordinates {(9, 55.46)(17, 57.36)(25, 58.81)(33, 58.85)};
            
        \legend{TranS4mer, Transformer}
        \end{axis}
        \end{tikzpicture}
    }
    }}\hspace{3mm}
\subfloat[\textbf{Training Speed $\uparrow$} TranS4mer is significantly faster than the Transformer baseline.\label{fig:transformer_speed}]{{
    \resizebox{0.28\textwidth}{!}{%
        \begin{tikzpicture}[yscale=0.9, font={\Large}]
        \begin{axis}[
            xlabel={\# Input Shots},
            ylabel={Samples/Sec.},
            xmin=8, xmax=34,
            xtick={9, 17, 25, 33},
            ymin=0.6, ymax=3.8,
            ytick={1.0, 2.0, 3.0},
            legend cell align={left},
            legend pos=north east
        ]
        \addplot[line width=.5mm, color=CornflowerBlue, mark=square, mark size=3pt]
            coordinates {(9, 3.40)(17, 2.91)(25, 2.57)(33, 2.24)};
        
        \addplot[line width=.5mm, color=Orange,mark=o, mark size=3pt]
            coordinates {(9, 2.86)(17, 1.93)(25, 1.27)(33, 0.83)};
            
        \legend{TranS4mer, Transformer}
        \end{axis}
        \end{tikzpicture}
    }
    }}\hspace{3mm}
\subfloat[\textbf{Memory Requirements $\downarrow$} TranS4mer requires significantly less GPU memory. \label{fig:transformer_memory}]{{
    \resizebox{0.28\textwidth}{!}{%
        \begin{tikzpicture}[yscale=0.9, font={\Large}]
        \begin{axis}[
            xlabel={\# Input Shots},
            ylabel={Memory (GB)},
            xmin=8, xmax=34,
            xtick={9, 17, 25, 33},
            ymin=4, ymax=49,
            legend cell align={left},
            legend pos=north west
        ]
        \addplot[line width=.5mm, color=CornflowerBlue, mark=square, mark size=3pt]
            coordinates {(9, 6.21)(17, 8.06)(25, 10.13)(33, 13.67)};
        
        \addplot[line width=.5mm, color=Orange,mark=o, mark size=3pt]
            coordinates {(9, 10.05)(17, 20.36)(25, 30.28)(33, 46.17)};
            
        \legend{TranS4mer, Transformer}
        \end{axis}
        \end{tikzpicture}
    }
    }}
\vspace{-0.1cm}
\caption{(a) Performance (AP), (b) Training Speed (Samples/Second), and (c) Memory Requirements (GigaBytes), and as a function of the number of input shots. Note that videos with more input shots denote longer input videos.}
\label{fig:comparison_transformer}

\end{figure*}
% \vspace{-2mm}

\begin{table*}[t]\centering
\subfloat[TranS4mer modules. \label{tab:architecture}]{
    \begin{tabular}{@{\extracolsep{0pt}}c c c}
        Intra-Shot & Inter-Shot & Ap ($\uparrow$)\\
        \toprule
        \cmark & \xmark & 57.80 \\
        \xmark & \cmark & 55.59 \\
        \cmark & \cmark & \bf 60.78 \\
        \bottomrule
    \end{tabular}
}\hspace{3mm}
\subfloat[S4 in different layers. \label{tab:s4_diff}]{
    \begin{tabular}{lc}
         S4 layers & AP ($\uparrow$) \\
         \toprule
         1-6 & 58.31 \\
         7-12 & 58.82 \\
         1-12 & \textbf{60.78} \\
         \bottomrule
    \end{tabular}
}\hspace{3mm}
\subfloat[S4 in every $k^{th}$ layer. \label{tab:s4_kth}]{
    \begin{tabular}{lc}
         S4 layers & AP ($\uparrow$) \\
         \toprule
         every 2nd & 59.82 \\
         every 4th & 58.01 \\
         all & \textbf{60.78} \\
         \bottomrule
    \end{tabular}
}\hspace{3mm}
\subfloat[Different S4 variants. \label{tab:s4_types}]{
    \begin{tabular}{lc}
         S4 layers & AP ($\uparrow$) \\
         \toprule
         S4 & 59.71 \\
         DS4 & 60.13 \\
         GS4 & \textbf{60.78} \\
         \bottomrule
    \end{tabular}
}
\vspace{-0.1cm}
\caption{Ablation studies with (a) TranS4mer modules, (b) (c) positioning, and (d) variants of S4 layers. Intra-shot and inter-shot modules are complementary, and TranS4mer with Gated S4 layers at every S4A block yields the best performance.}
\vspace{-0.1cm}
\label{tab:ablation}
\end{table*}

\subsection{Long-range Video Classification}
In addition to move scene detection, we also experiment with other tasks including long-form movie clip classification in the LVU~\cite{wu2021towards} dataset, and long-range procedural activity classification on Breakfast~\cite{kuehne2014language} and COIN~\cite{tang2019coin} datasets. LVU contains $~30K$ videos from $~3K$ movies, Breakfast has 1,712 videos of 10 cooking activities, and COIN contains 11,827 videos of 180 different procedural activities. Videos of each dataset are $~2-3$ minutes long. We present the results in Table \ref{tab:video_classification}. On LVU, TranS4mer achieves the best performance in 5 out of 7 tasks (Table \ref{tab:video_classification}(a)). Moreover, TranS4mer also achieves the best performance on Breakfast (Table \ref{tab:video_classification}(b)) and the second-best performance on COIN (Table \ref{tab:video_classification}(c)) while using significantly fewer pretraining samples than the state-of-the-art Distant Supervision~\cite{lin2022learning} approach. These results indicate that TranS4mer can be effectively applied to other domains and tasks. %such as long-form video understanding and procedural activity classification. 

\subsection{Ablation Studies}

\textbf{Temporal Extent Ablation.} To verify our hypothesis that long-range temporal context is critical for accurate movie scene detection, we experiment with video inputs of different lengths (i.e., video inputs containing 9, 17, 25, and 33 contextual shots) while predicting the boundary for the middle shot. Note that videos with more input shots denote longer input videos. We present these results in Figure~\ref{fig:transformer_performance}. First, we notice that a short temporal window of 9 contextual shots produces a suboptimal performance of \textbf{53.29\%} AP. Increasing the temporal extent to 25 contextual shots increases the performance to \textbf{60.78\%} AP (\textbf{+7.49\%} boost). Adding more contextual shots does not yield any significant improvements. These results verify the importance of long-range temporal modeling for movie scene detection. 

 \textbf{Detailed Comparison with Transformer.} We next present a detailed comparison of our model with the Transformer baseline~\cite{dosovitskiy2020image}. In particular, we experiment with videos that contain a different number of input shots and plot the accuracy in AP as a function of the number of input shots (see Figure~\ref{fig:comparison_transformer}). Although the Transformer baseline performs better for shorter video inputs, our TranS4mer model achieves much better performance using longer videos. Specifically, TranS4mer yields \textbf{+2.07}\% larger AP than the transformer model when using the longest video inputs. This  suggests that TranS4mer is more effective at capturing long-range dependencies for the movie scene detection task. Moreover, we also illustrate the training speed (Samples/Second) and the memory requirement (GigaBytes) as a function of the number of input shots (see Figures~\ref{fig:transformer_speed} and ~\ref{fig:transformer_memory}). We observe that as we increase the video length, the training speed of the Transformer baseline drops drastically, and the memory requirement grows rapidly. Overall, our proposed TranS4mer model is $\textbf{2.7}\times$ faster and requires $\textbf{3.38}\times$ less GPU memory than the Transformer baseline while using the longest video inputs. 

 \textbf{Ablation with Intra-Shot and Inter-Shot Modules.} In Table~\ref{tab:architecture}, we analyze the significance of our (i) self-attention-based intra-shot module and (ii) state-space-based inter-shot module. We observe that removing the inter-shot module drops the performance by \textbf{2.98\%} AP, whereas discarding the intra-shot module decreases the performance by \textbf{5.19\%} AP. Therefore, based on these results, we can conclude that both the intra-shot and inter-shot modules are essential for good movie scene detection performance. We also note that they are complementary to each other, which aligns with our intuition of the intra-shot module being responsible for short-range representations while the inter-shot module is dedicated to long-range temporal reasoning.

 \textbf{Ablation with S4 Layers.}
We experiment with injecting our proposed S4A block into different encoder layers (Table~\ref{tab:s4_diff}). We also evaluate a sparser TranS4mer variant with our S4A block injected into every $k^{th}$ layer of the network (Table~\ref{tab:s4_kth}). Lastly, we experiment with different S4 variants (i.e., Diagonal S4 (DS4)~\cite{gupta2022diagonal}, Gated S4 (GS4)~\cite{mehta2022long}) (Table~\ref{tab:s4_types}). Our results suggest that using S4A with Gated S4 in every layer of the network leads to the best accuracy.

\section{Conclusion}

In this work, we propose an efficient movie scene segmentation method which combines the strengths of self-attention for short-range modeling and state-space models for long-range modeling. Our method, named TranS4mer, achieves state-of-the-art results in three different movie scene segmentation datasets (MovieNet, BBC, and OVSD). Moreover, our model is significantly faster and requires less GPU memory than previous approaches and the standard vision transformer models. In future, we will extend our model to other long-range video understanding and multi-modal tasks such as movie summarization, movie trailer generation, and natural language grounding in long videos.

%%%%%%%%% REFERENCES
{\small
\bibliographystyle{ieee_fullname}
\bibliography{egbib}
}

\end{document}